%% file: main.tex
\documentclass{article}
\usepackage{dsfont}
\usepackage{spconf,graphicx}
\usepackage{booktabs}
\usepackage{amsmath}

\usepackage{amssymb}
\usepackage{amsfonts}
\usepackage{pifont}
\def\cmark{\ding{51}} 
\def\xmark{\ding{55}}
\usepackage{upgreek}
\usepackage{makecell}
\usepackage{array}
\usepackage{units}
\usepackage{multirow}
\usepackage{makecell}
\usepackage{tabularx}
\usepackage{url}
\usepackage{cite}
\usepackage{tikz}
\usepackage{pgfplots}
\usepgfplotslibrary{colorbrewer}
\pgfplotsset{cycle list/Set1-9}
\tikzset{every picture/.style={line width=1pt}}
\usetikzlibrary{patterns}
\usepgfplotslibrary{fillbetween}
\pgfplotsset{
  tick label style = {font=\sansmath\sffamily\scriptsize},
  every axis label = {font=\sansmath\sffamily\scriptsize},
  y label style={at={(0.05,0.5)}},
  legend style = {font=\sansmath\sffamily\scriptsize},
  label style = {font=\sansmath\sffamily\scriptsize},
}
\pgfplotsset{compat=1.3} %
\newlength{\Oldarrayrulewidth}
\newcommand{\Cline}[2]{%
  \noalign{\global\setlength{\Oldarrayrulewidth}{\arrayrulewidth}}%
  \noalign{\global\setlength{\arrayrulewidth}{#1}}\cline{#2}%
  \noalign{\global\setlength{\arrayrulewidth}{\Oldarrayrulewidth}}}
\DeclareMathAlphabet\mathbfcal{OMS}{cmsy}{b}{n}

\DeclareMathOperator*{\argmin}{\arg\!\min}

\title{A Novel Self-Supervised Cross-Modal Image Retrieval\\Method In Remote Sensing}
\name{Gencer Sumbul, Markus M\"{u}ller, Beg\"{u}m Demir}
\address{Faculty of Electrical Engineering and Computer Science, Technische Universit\"at Berlin, Germany}
\global\hyphenpenalty=1
\begin{document}
%
\maketitle
\begin{abstract}
Due to the availability of multi-modal remote sensing (RS) image archives, one of the most important research topics is the development of cross-modal RS image retrieval (CM-RSIR) methods that search semantically similar images across different modalities. Existing CM-RSIR methods require the availability of a high quality and quantity of annotated training images. The collection of a sufficient number of reliable labeled images is time consuming, complex and costly in operational scenarios, and can significantly affect the final accuracy of CM-RSIR. In this paper, we introduce a novel self-supervised CM-RSIR method that aims to: i) model mutual-information between different modalities in a self-supervised manner; ii) retain the distributions of modal-specific feature spaces similar to each other; and iii) define the most similar images within each modality without requiring any annotated training image. To this end, we propose a novel objective including three loss functions that simultaneously: i) maximize mutual information of different modalities for inter-modal similarity preservation; ii) minimize the angular distance of multi-modal image tuples for the elimination of inter-modal discrepancies; and iii) increase cosine similarity of the most similar images within each modality for the characterization of intra-modal similarities. Experimental results show the effectiveness of the proposed method compared to state-of-the-art methods. The code of the proposed method is publicly available at \url{https://git.tu-berlin.de/rsim/SS-CM-RSIR}.
\end{abstract}
%
\section{Introduction}
Recent years have witnessed the availability of ever-growing multi-modal remote sensing (RS) image archives. Accordingly, one of the most important research topics in RS is the development of cross-modal RS image retrieval (CM-RSIR) methods that search semantically similar images across different modalities. Although uni-modal RS image retrieval problems have been excessively studied (the reader is referred to~\cite{Sumbul:2021} for a detailed survey), a few CM-RSIR methods have been introduced in RS~\cite{Chaudhuri:2020, Xiong:2020, Sun:2021}. When image representation learning is employed for CM-RSIR, it is crucial to characterize intra- and inter-modal similarities, while eliminating inter-modal discrepancies due to semantic gap among different modalities. These issues have been separately addressed by the existing CM-RSIR methods. In~\cite{Chaudhuri:2020}, a deep neural network based framework for cross-modal retrieval (CMIR-NET) is introduced to learn an unified representation space over the modal-specific features. To this end, CMIR-NET employs a two-stage training procedure, in which the first stage is dedicated to supervised modal-specific feature learning with cross-entropy objective for intra-modal similarity preserving. The second stage is employed for learning the unified representation space by using mean squared error, supervised cross-entropy and reconstruction losses among different modalities that preserves inter-modal similarities. In~\cite{Xiong:2020}, a deep cross-modality hashing network (DCHMN) is proposed to eliminate inter-modal discrepancy while characterizing intra-modal similarities. To this end, DCHMN employs supervised triplet loss function on the image representations of each modality for intra-modal similarity preserving. To eliminate the inter-modal discrepancy, DCHMN exploits a convolutional neural network (CNN) shared by different modalities that accepts randomly selected image bands from each modality. In~\cite{Sun:2021}, a multi-sensor fusion and explicit semantic preserving-based deep hashing method (MsEspH) is introduced to apply fusion with synthetically generated images that can reduce the inter-modal discrepancy among different modalities. MsEspH also employs supervised cross-entropy objective on each modality for the characterization of intra-modal similarities. The existing CM-RSIR systems require the availability of training images annotated by land-use land-cover classes. However, either the availability of labelled multi-modal RS image archives is limited or collecting reliable RS image labels can be time consuming and expensive due to labeling costs. In greater details, these methods partially consider intra- and inter-modal similarity preservation and inter-modal discrepancy elimination that are not simultaneously achieved yet. To address the above mentioned issues, we introduce a novel self-supervised CM-RSIR method. The proposed method is independent from the type of RS image modalities (which are considered for CM-RSIR tasks), and thus can be applied to different multi-modal RS image archives.
\section{Proposed Self-Supervised\\CM-RSIR Method}
Let $\{\mathcal{X}^1,\ldots,\mathcal{X}^N\}$ be a set of $N$ image archives, each of which is associated with a different image modality (acquired by a different sensor). An archive $\mathcal{X}^j=\{\boldsymbol{x}_i^j\}_{i=1}^M$ includes $M$ images, where $\boldsymbol{x}_i^j$ is the $i$th RS image in the $j$th image archive and $(\boldsymbol{x}_i^j)_{j=1}^N$ is the $i$th multi-modal image tuple that includes $N$ images acquired by different sensors on the same geographical area. A CM-RSIR task can be formulated as finding images from the image archive $\mathcal{X}^j$ that are similar to a given query from the image archive $\mathcal{X}^k$ while $j \neq k$. The proposed method aims to simultaneously preserve intra- and inter-modal similarities and eliminate inter-modal discrepancies without requiring annotated training images. This is achieved by a novel self-supervised objective that includes three loss functions to: i) maximize mutual information of different modalities for inter-modal similarity preservation; ii) minimize the angular distance of multi-modal image tuples for the elimination of inter-modal discrepancies; and iii) maximize the cosine similarity of the most similar images within each modality for intra-modal similarity preservation. To this end, the proposed objective can be integrated in any deep neural network (DNN), which embodies one CNN backbone for each modality followed by fully connected (FC) layers as the common cross-modal encoder for different modalities. Let $f^j : \{\boldsymbol{x}_i^j\}_{i=1}^M \mapsto \mathbb{R}^{a}$ be a CNN backbone that characterizes the modal-specific image features $\{\boldsymbol{y}_i^j\}_{i=1}^M$ having the size of $a$ for the $j$th modality. Let $g : \{(\boldsymbol{y}_i^j, \boldsymbol{y}_i^k)\}_{i=1}^M \mapsto \mathbb{R}^{b}$ be a common encoder that maps the modal-specific image features into cross-modal embeddings $\{(\boldsymbol{z}_i^j, \boldsymbol{z}_i^k)\}_{i=1}^M$, where $b$ is the embedding size. After learning the parameters of $f^j$, $f^k$ and $g$ based on the proposed objective, $g$ can be used to extract the embedding of an image from any modality that is utilized to perform CM-RSIR. Fig. 1 shows a general overview of the proposed method, which is explained in detail in the following subsections.
\subsection{Inter-Modal Similarity Preservation}
To preserve inter-modal similarities, inspired by the information theory, we define the capability of preserving inter-modal similarities in the considered DNN as the mutual information between the image embeddings $\{(\boldsymbol{z}_i^j, \boldsymbol{z}_i^k)\}_{i=1}^M$. Let $I(\boldsymbol{z}_i^j; \boldsymbol{z}_i^k)$ be the mutual information between $j$th and $k$th embeddings of the $i$th tuple $(\boldsymbol{x}_i^j, \boldsymbol{x}_i^k)$. If $I(\boldsymbol{z}_i^j; \boldsymbol{z}_i^k)$ is high, $g$ accurately models the shared semantic content between $j$th and $k$th modalities (which leads to preservation of inter-modal similarities) and vice versa. However, directly maximizing the mutual information is not feasible since the true posterior distributions of $\boldsymbol{z}_i^j$ and $\boldsymbol{z}_i^k$ ($P(\boldsymbol{z}_i^j | \boldsymbol{z}_i^k)$ and $P(\boldsymbol{z}_i^k | \boldsymbol{z}_i^j)$) are unknown \cite{Hoang:2022}. To this end, we utilize the normalized temperature-scaled cross entropy (NT-Xent) objective $\mathcal{L}_{\text{NT-Xent}}$ in a self-supervised manner. In this way, image embeddings are characterized by maximizing agreement between the multiple views of a shared context compared to the negative samples \cite{Bachman:2019}. It is worth noting that NT-Xent objective applied to the features of these multiple views is known to be a lower bound to the mutual information of their features \cite{Poole:2019} as:
\begin{equation}
    I(v_1; v_2) \geq log(K) - \mathcal{L}_{\text{NT-Xent}}(v_1, v_2),  
\end{equation}
where $K$ is the number of negative samples and $\{v_1, v_2\}$ represents the features of two multiple views. In the existing applications of $\mathcal{L}_{\text{NT-Xent}}$, two different views are obtained via data augmentation techniques applied on the same image. In this paper, we consider the images of the $i$th image tuple $(\boldsymbol{x}_i^j, \boldsymbol{x}_i^k)$ as the multiple views of the same geographical area and the images of other tuples as the negative samples. To this end, for each training mini-batch $\mathbfcal{B}$ that includes $T$ image tuples, we define our mutual information maximization loss function $\mathcal{L}_{\text{MIM}}$ for different modalities based on the NT-Xent objective as follows:
\begin{equation}
\begin{aligned}
\ell^i(j,k)  = -\text{log}(\frac{e^{S(\boldsymbol{z}_i^j, \boldsymbol{z}_i^k)/\tau}}{\sum_{q=1}^{T}\mathds{1}_{[q \neq i]}e^{S(\boldsymbol{z}_i^j, \boldsymbol{z}_q^k)/\tau}}),\\
\mathcal{L}_{\text{MIM}}(\{(\boldsymbol{z}_i^j, \boldsymbol{z}_i^k)\}_{i=1}^T)\! =\! \frac{1}{2T}\sum_{i=1}^T \ell^i(j,k)\! +\! \ell^i(k,j),\\
\end{aligned}
\end{equation}
where $S(\cdot, \cdot)$ measures cosine similarity, $\mathds{1}$ is the indicator function, $\tau$ is the temperature parameter and $K\!=\!T\!-\!1$. Training the considered DNN with $\mathcal{L}_{\text{MIM}}$ leads to maximizing the similarity of images acquired on the same geographical area from different modalities on the common cross-modal embedding space, which is defined by $g$. Based on (1), decreasing $\mathcal{L}_{\text{MIM}}$ over cross-modal embeddings $\boldsymbol{z}_i^j, \boldsymbol{z}_i^k$ leads to increasing the mutual information between them $I(\boldsymbol{z}_i^j; \boldsymbol{z}_i^k)$ for the $i$th image tuple $(\boldsymbol{x}_i^j, \boldsymbol{x}_i^k)$ in the archive.
\input{model_fig2}
\subsection{Inter-Modal Discrepancy Elimination}
To eliminate inter-modal discrepancies, the proposed method keeps the distributions of modal-specific image features similar to each other on different feature spaces. In the framework of CM-RSIR, the characteristics of RS images acquired by different sensors on the same geographical area can be significantly different due to profound differences in acquisition conditions, sensor characteristics and considered electromagnetic spectrum. Once modal-specific backbones are used for the feature learning of each modality, inter-modal discrepancies can be encoded by them that also leads to inaccurate inter-modal similarity preservation. One could also utilize a common backbone for different modalities to address this issue. However, this could lead to inaccurate characterization of image features in each modality. Accordingly, for inter-modal discrepancy elimination occurred due to using modal-specific backbones, we define the inter-modal discrepancy elimination loss function $\mathcal{L}_{\text{MDE}}$ over modal-specific backbones for each training mini-batch $\mathbfcal{B}$ as follows:
\begin{equation}
    \mathcal{L}_{\text{MDE}}(\{(\boldsymbol{y}_i^j, \boldsymbol{y}_i^k)\}_{i=1}^T)\!=\! -\frac{1}{T} \sum_{i=1}^T \text{log}(1+e^{S(\boldsymbol{y}_i^j, \boldsymbol{y}_i^k)}).
\end{equation}
This loss function enforces to have a small angular distance between the modal-specific image features $\boldsymbol{y}_i^j$ and $\boldsymbol{y}_i^k$ of the $j$th and $k$th modalities although they are obtained from different backbones. This leads to modelling similar distributions of modal-specific image features on different feature spaces of $f^j$ and $f^k$. Accordingly, the proposed method prevents the modal-specific backbones from characterizing the inter-modal discrepancies. 
\subsection{Intra-Modal Similarity Preservation}
To preserve intra-modal similarities, the proposed method defines the most similar images in each modality and further decrease their distance in the corresponding feature space. In the framework of CM-RSIR, finding images similar to each other for a given query depends on the effectiveness of the image representations in terms of their capability to encode intra-modal similarities. Since the proposed method does not require the availability of annotated training images, it is not suitable to utilize supervised triplet or cross-entropy loss functions for intra-modal similarity preservation unlike the existing CM-RSIR methods. To this end, we utilize each modal-specific feature space to measure the similarities of images from the corresponding modality. This forms the basis for defining semantically similar images within each modality. Once the most similar images within each modality are defined, their closeness in the modal-specific feature space can be further increased. This leads to preserving intra-modal similarities for all modalities. To this end, for a training mini-batch $\mathbfcal{B}$, we define the intra-modal similarity preservation loss function $\mathcal{L}_{\text{MSP}}$ over modal-specific feature spaces as follows:
\begin{equation}
\begin{aligned}
s_{ij} = S(\boldsymbol{y}_i^j,\! \argmin\limits_{\boldsymbol{y}_k^j, k\neq i} D(\boldsymbol{y}_i^j, \boldsymbol{y}_k^j))\\
    \mathcal{L}_{\text{MSP}}(\{(\boldsymbol{y}_i^j, \boldsymbol{y}_i^k)\}_{i=1}^T)\! =\! \frac{-1}{2T} \!\sum_{i}^{T}\!s_{ij} + s_{ik},
\end{aligned}
\end{equation}
where $D(\cdot,\cdot)$ is the image similarity measure within one modality, defined as the Euclidean distance between two image features in a modal-specific feature space. $\mathcal{L}_{\text{MSP}}$ enforces to maximize the cosine similarity of the most similar image pairs within each modality, and thus intra-modal similarity preservation for all modalities. 

By combining (2), (3) and (4), the proposed objective of our method is obtained as $\mathcal{L} = \mathcal{L}_{\text{MIM}} + \alpha\mathcal{L}_{\text{MDE}} + \beta\mathcal{L}_{\text{MSP}}$, where $\alpha$ and $\beta$ are the weighting parameters of $\mathcal{L}_{\text{MDE}}$ and $\mathcal{L}_{\text{MSP}}$, respectively. Increasing the importance of $\mathcal{L}_{\text{MDE}}$ and $\mathcal{L}_{\text{MSP}}$ more than $\mathcal{L}_{\text{MIM}}$ on an early stage of training can lead to sub-optimal solutions. Accordingly, we increase $\alpha$ and $\beta$ throughout epochs with the exponential growth scheduling strategy. 
\section{Experimental Results}
\input{vis_res_fig}
Experiments were conducted on the BigEarthNet-MM archive \cite{BigEarthNetMM} that includes 590,326 multi-modal image pairs. Each pair in BigEarthNet-MM includes one Sentinel-1 (denoted as S1) SAR image and one Sentinel-2 (denoted as S2) multispectral image acquired on the same geographical area. Each pair is associated with one or more class labels (i.e., multi-labels) based on the 19 class nomenclature. To perform experiments, we first selected the 14,832 BigEarthNet-MM image pairs acquired over Serbia during summer season and then divided them into training (52\%), validation (24\%) and test (24\%) sets. The training set was used to perform training and to select query images, while images were retrieved from the test set. In the experiments, we utilized the DenseNet model \cite{Huang:2017} at the depth of 121 as the CNN backbone, while one FC layer with the size of 128 was used for common cross-modal encoder. The initial values of $\alpha$ and $\beta$ were set to $10^{-4}$, which exponentially increased to 1 until the end of training. We trained our method for 100 epochs with the mini-bath size of 256 by using the Adam optimizer. To assess the effectiveness of the proposed method, we compared it with state-of-the-art fully-supervised (S2MC~\cite{Li:2020}, deep-SM~\cite{Wei:2017}, DSCMR~\cite{Zhen:2019}), unsupervised (DCCA~\cite{Andrew:2013}) and self-supervised (SimCLR~\cite{Chen:2019}, DUCH~\cite{Mikriukov:2022}) methods. Hash code learning related loss functions used in DUCH were eliminated for a fair comparison. All the methods were trained by using the same CNN backbone architecture under the same experimental setup. For the NT-Xent objectives used in the experiments, the temperature parameter $\tau$ was set to 0.2. In the experiments, the results of S2$\to$S1 and S1$\to$S2 retrieval tasks are given in terms of $F_1$ score and normalized discounted cumulative gain (NDCG). The retrieval performance was assessed on top-8 retrieved images. 
\input{exp_result_table} 

Table 1 shows the corresponding cross-modal retrieval performances. By assessing the table, one can observe that the proposed method leads to the highest scores under both metrics and retrieval tasks compared to the unsupervised and self-supervised methods. This shows that the proposed method accurately addresses the intra- and inter-modal similarities while eliminating inter-modal discrepancies without requiring any training image label. One can also observe from the table that the proposed method provides the highest scores compared to all supervised methods except DSCMR. Compared to DSCMR, it achieves similar NDCG and $F_1$ scores for S1$\to$S2 and S2$\to$S1 retrieval tasks, respectively. However, DSCMR provides 1\% higher metric scores on average than the proposed method at the cost of the training sample requirements (DSCMR utilized 21,367 labeled images to reach this performance). Fig. 2 shows an example of S2 images retrieved by the best performing three methods when the S1 query image contains \textit{Urban fabric}, \textit{Arable land} and \textit{Land principally occupied by agriculture} classes. By assessing the figure, one can see that all the images retrieved by the proposed method contain all these classes. However, the retrieved images by self-supervised DUCH method and supervised DSCMR method contain \textit{Urban fabric} class at only some of the retrieved images. One can observe from the figure that the proposed method applies CM-RSIR more accurately than DUCH and as effective as DSCMR.
\section{Conclusion}
In this paper, we have introduced a novel self-supervised CM-RSIR method. The effectiveness of our method relies on efficiently preserving intra- and inter-modal similarities and eliminating inter-modal discrepancies without requiring any annotated training image. This is achieved by a novel objective that leads to: i) modelling mutual-information across different modalities in a self-supervised manner; ii) retaining the distributions of modal-specific feature spaces similar to each other; and iii) defining the most similar images for each modality. Experimental results show the success of the proposed method compared to state-of-art unsupervised, self-supervised and fully supervised methods in the framework of CM-RSIR. As a future work, we plan to integrate our method to cross-modal text-image retrieval problems to search for RS images using a query text sentence. 
\section{Acknowledgements}
This work is funded by the European Research Council (ERC) through the ERC-2017-STG BigEarth Project under Grant 759764.
\vfill
\pagebreak

\bibliographystyle{IEEEbib}
\bibliography{defs,refs}

\end{document}

%% file: model_fig2.tex
\begin{figure}[t]
    \centering

\tikzset{every picture/.style={line width=0.75pt}} 

\begin{tikzpicture}[x=0.75pt,y=0.75pt,yscale=-1,xscale=1]

\draw (45,46) node  {\includegraphics[width=52.5pt,height=52.5pt]{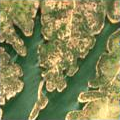}};
\draw  [fill={rgb, 255:red, 65; green, 117; blue, 5 }  ,fill opacity=0.5 ] (102,11) -- (157.02,27.51) -- (157.02,64.49) -- (102,81) -- cycle ;
\draw (45.5,151.25) node  {\includegraphics[width=52.5pt,height=52.5pt]{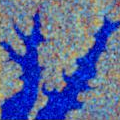}};
\draw  [fill={rgb, 255:red, 155; green, 155; blue, 155 }  ,fill opacity=0.5 ] (216.1,12.61) .. controls (216.1,10.62) and (217.72,9) .. (219.71,9) -- (230.55,9) .. controls (232.54,9) and (234.16,10.62) .. (234.16,12.61) -- (234.16,77.6) .. controls (234.16,79.6) and (232.54,81.21) .. (230.55,81.21) -- (219.71,81.21) .. controls (217.72,81.21) and (216.1,79.6) .. (216.1,77.6) -- cycle ;
\draw  [fill={rgb, 255:red, 208; green, 2; blue, 27 }  ,fill opacity=0.5 ] (218.6,18.28) .. controls (218.6,14.81) and (221.41,12) .. (224.88,12) .. controls (228.35,12) and (231.16,14.81) .. (231.16,18.28) .. controls (231.16,21.75) and (228.35,24.56) .. (224.88,24.56) .. controls (221.41,24.56) and (218.6,21.75) .. (218.6,18.28) -- cycle ;
\draw  [fill={rgb, 255:red, 208; green, 2; blue, 27 }  ,fill opacity=0.5 ] (218.1,34.53) .. controls (218.1,30.92) and (221.02,28) .. (224.63,28) .. controls (228.23,28) and (231.16,30.92) .. (231.16,34.53) .. controls (231.16,38.13) and (228.23,41.06) .. (224.63,41.06) .. controls (221.02,41.06) and (218.1,38.13) .. (218.1,34.53) -- cycle ;
\draw  [fill={rgb, 255:red, 208; green, 2; blue, 27 }  ,fill opacity=0.5 ] (218.1,71.53) .. controls (218.1,67.92) and (221.02,65) .. (224.63,65) .. controls (228.23,65) and (231.16,67.92) .. (231.16,71.53) .. controls (231.16,75.13) and (228.23,78.06) .. (224.63,78.06) .. controls (221.02,78.06) and (218.1,75.13) .. (218.1,71.53) -- cycle ;
\draw  [dash pattern={on 0.84pt off 2.51pt}]  (224.51,48.5) -- (224.66,59.64) ;
\draw    (80.56,45.64) -- (98.15,45.64) ;
\draw [shift={(101.15,45.64)}, rotate = 180] [fill={rgb, 255:red, 0; green, 0; blue, 0 }  ][line width=0.08]  [draw opacity=0] (7.14,-3.43) -- (0,0) -- (7.14,3.43) -- cycle    ;
\draw    (156.68,44.65) -- (212.16,44.64) ;
\draw [shift={(215.16,44.64)}, rotate = 179.99] [fill={rgb, 255:red, 0; green, 0; blue, 0 }  ][line width=0.08]  [draw opacity=0] (7.14,-3.43) -- (0,0) -- (7.14,3.43) -- cycle    ;
\draw  [fill={rgb, 255:red, 74; green, 144; blue, 226 }  ,fill opacity=0.5 ] (102.5,115.75) -- (157.52,132.26) -- (157.52,169.24) -- (102.5,185.75) -- cycle ;
\draw  [fill={rgb, 255:red, 155; green, 155; blue, 155 }  ,fill opacity=0.5 ] (216.6,117.36) .. controls (216.6,115.37) and (218.22,113.75) .. (220.21,113.75) -- (231.05,113.75) .. controls (233.04,113.75) and (234.66,115.37) .. (234.66,117.36) -- (234.66,182.35) .. controls (234.66,184.35) and (233.04,185.96) .. (231.05,185.96) -- (220.21,185.96) .. controls (218.22,185.96) and (216.6,184.35) .. (216.6,182.35) -- cycle ;
\draw  [fill={rgb, 255:red, 208; green, 2; blue, 27 }  ,fill opacity=0.5 ] (219.1,123.03) .. controls (219.1,119.56) and (221.91,116.75) .. (225.38,116.75) .. controls (228.85,116.75) and (231.66,119.56) .. (231.66,123.03) .. controls (231.66,126.5) and (228.85,129.31) .. (225.38,129.31) .. controls (221.91,129.31) and (219.1,126.5) .. (219.1,123.03) -- cycle ;
\draw  [fill={rgb, 255:red, 208; green, 2; blue, 27 }  ,fill opacity=0.5 ] (218.6,139.28) .. controls (218.6,135.67) and (221.52,132.75) .. (225.13,132.75) .. controls (228.73,132.75) and (231.66,135.67) .. (231.66,139.28) .. controls (231.66,142.88) and (228.73,145.81) .. (225.13,145.81) .. controls (221.52,145.81) and (218.6,142.88) .. (218.6,139.28) -- cycle ;
\draw  [fill={rgb, 255:red, 208; green, 2; blue, 27 }  ,fill opacity=0.5 ] (218.6,176.28) .. controls (218.6,172.67) and (221.52,169.75) .. (225.13,169.75) .. controls (228.73,169.75) and (231.66,172.67) .. (231.66,176.28) .. controls (231.66,179.88) and (228.73,182.81) .. (225.13,182.81) .. controls (221.52,182.81) and (218.6,179.88) .. (218.6,176.28) -- cycle ;
\draw  [dash pattern={on 0.84pt off 2.51pt}]  (225.01,153.25) -- (225.16,164.39) ;
\draw    (81.06,150.39) -- (98.65,150.39) ;
\draw [shift={(101.65,150.39)}, rotate = 180] [fill={rgb, 255:red, 0; green, 0; blue, 0 }  ][line width=0.08]  [draw opacity=0] (7.14,-3.43) -- (0,0) -- (7.14,3.43) -- cycle    ;
\draw    (157.98,149.8) -- (212.66,149.41) ;
\draw [shift={(215.66,149.39)}, rotate = 179.59] [fill={rgb, 255:red, 0; green, 0; blue, 0 }  ][line width=0.08]  [draw opacity=0] (7.14,-3.43) -- (0,0) -- (7.14,3.43) -- cycle    ;
\draw    (179.56,44.64) -- (179.5,82.25) ;
\draw [shift={(179.5,85.25)}, rotate = 270.08] [fill={rgb, 255:red, 0; green, 0; blue, 0 }  ][line width=0.08]  [draw opacity=0] (7.14,-3.43) -- (0,0) -- (7.14,3.43) -- cycle    ;
\draw    (234.18,44) -- (279.62,43.85) -- (279.97,82.25) ;
\draw [shift={(280,85.25)}, rotate = 269.47] [fill={rgb, 255:red, 0; green, 0; blue, 0 }  ][line width=0.08]  [draw opacity=0] (7.14,-3.43) -- (0,0) -- (7.14,3.43) -- cycle    ;
\draw  [dash pattern={on 4.5pt off 4.5pt}] (205.42,10.65) .. controls (205.42,6.29) and (208.96,2.75) .. (213.32,2.75) -- (237.01,2.75) .. controls (241.37,2.75) and (244.9,6.29) .. (244.9,10.65) -- (244.9,185.98) .. controls (244.9,190.34) and (241.37,193.88) .. (237.01,193.88) -- (213.32,193.88) .. controls (208.96,193.88) and (205.42,190.34) .. (205.42,185.98) -- cycle ;
\draw    (225.03,108.5) -- (225.22,86.5) ;
\draw [shift={(225.25,83.5)}, rotate = 90.51] [fill={rgb, 255:red, 0; green, 0; blue, 0 }  ][line width=0.08]  [draw opacity=0] (7.14,-3.43) -- (0,0) -- (7.14,3.43) -- cycle    ;
\draw [shift={(225,111.5)}, rotate = 270.51] [fill={rgb, 255:red, 0; green, 0; blue, 0 }  ][line width=0.08]  [draw opacity=0] (7.14,-3.43) -- (0,0) -- (7.14,3.43) -- cycle    ;
\draw    (180,149.5) -- (180.23,111.25) ;
\draw [shift={(180.25,108.25)}, rotate = 90.35] [fill={rgb, 255:red, 0; green, 0; blue, 0 }  ][line width=0.08]  [draw opacity=0] (7.14,-3.43) -- (0,0) -- (7.14,3.43) -- cycle    ;
\draw    (235.9,150.35) -- (280.38,149.94) -- (280.49,111.25) ;
\draw [shift={(280.5,108.25)}, rotate = 90.16] [fill={rgb, 255:red, 0; green, 0; blue, 0 }  ][line width=0.08]  [draw opacity=0] (7.14,-3.43) -- (0,0) -- (7.14,3.43) -- cycle    ;

\draw (144.4,90.43) node [anchor=north west][inner sep=0.75pt]  [font=\footnotesize]  {$\mathcal{L}_{\text{MDE}} ,\mathcal{L}_{\text{MSP}}$};
\draw (265.95,90.43) node [anchor=north west][inner sep=0.75pt]  [font=\footnotesize]  {$\mathcal{L}_{\text{MIM}}$};
\draw (37.1,82.73) node [anchor=north west][inner sep=0.75pt]  [font=\footnotesize]  {$\boldsymbol{x}_{i}^{j}$};
\draw (39.6,187.48) node [anchor=north west][inner sep=0.75pt]  [font=\footnotesize]  {$\boldsymbol{x}_{i}^{k}$};
\draw (126.3,74.73) node [anchor=north west][inner sep=0.75pt]  [font=\footnotesize]  {$f^{j}$};
\draw (126.8,179.08) node [anchor=north west][inner sep=0.75pt]  [font=\footnotesize]  {$f^{k}$};
\draw (116,40) node [anchor=north west][inner sep=0.75pt]  [font=\footnotesize] [align=left] {CNN};
\draw (116.1,144.75) node [anchor=north west][inner sep=0.75pt]  [font=\footnotesize] [align=left] {CNN};
\draw (221.2,193.9) node [anchor=north west][inner sep=0.75pt]  [font=\footnotesize]  {$g$};
\draw (160.3,25.63) node [anchor=north west][inner sep=0.75pt]  [font=\footnotesize]  {$\boldsymbol{y}_{i}^{j}$};
\draw (159.6,131.08) node [anchor=north west][inner sep=0.75pt]  [font=\footnotesize]  {$\boldsymbol{y}_{i}^{k}$};
\draw (246.21,26.02) node [anchor=north west][inner sep=0.75pt]  [font=\footnotesize]  {$\boldsymbol{z}_{i}^{j}$};
\draw (246.71,131.15) node [anchor=north west][inner sep=0.75pt]  [font=\footnotesize]  {$\boldsymbol{z}_{i}^{k}$};

\end{tikzpicture}

    \caption{Illustration of the proposed self-supervised cross-modal content-based image retrieval method.}
\label{fig:model}
\end{figure}

%% file: vis_res_fig.tex
\begin{figure*}[ht!]
    \newcommand{\figwidth}{0.11\linewidth}
    \newcommand{\figheight}{0.65in}
    \renewcommand{\fboxsep}{0pt}
    \newcommand{\rfig}[3]{vis_result/#1/#2/#3.png}
    \newcommand{\qfig}[1]{vis_result/#1/query.png}
    \newcommand{\efig}{vis_result/empty.png}
    \centering
     \begin{minipage}[t]{\figwidth}
        \centering
        \centerline{}\medskip
        \centerline{\includegraphics[height=\figheight]{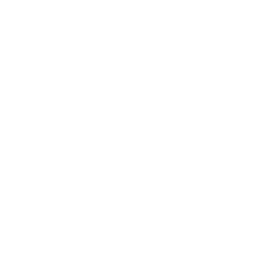}}
    \end{minipage}
     \begin{minipage}[t]{\figwidth}
        \centering
        \centerline{1\textsuperscript{st}}\medskip
        \centerline{\includegraphics[height=\figheight]{\rfig{S1A_IW_GRDH_1SDV_20170802T163350_34TCR_74_25}{DUCH}{1}}} 
    \end{minipage}
          \begin{minipage}[t]{\figwidth}
        \centering
        \centerline{2\textsuperscript{nd}}\medskip
        \centerline{\includegraphics[height=\figheight]{\rfig{S1A_IW_GRDH_1SDV_20170802T163350_34TCR_74_25}{DUCH}{2}}} 
    \end{minipage}    
          \begin{minipage}[t]{\figwidth}
        \centering
        \centerline{3\textsuperscript{rd}}\medskip
        \centerline{\includegraphics[height=\figheight]{\rfig{S1A_IW_GRDH_1SDV_20170802T163350_34TCR_74_25}{DUCH}{3}}} 
    \end{minipage}
          \begin{minipage}[t]{\figwidth}
        \centering
        \centerline{4\textsuperscript{th}}\medskip
        \centerline{\includegraphics[height=\figheight]{\rfig{S1A_IW_GRDH_1SDV_20170802T163350_34TCR_74_25}{DUCH}{4}}} 
        \vspace{-0.05in}\centerline{(b)}\medskip\vspace{-0.06in}
    \end{minipage}
          \begin{minipage}[t]{\figwidth}
        \centering
        \centerline{5\textsuperscript{th}}\medskip
        \centerline{\includegraphics[height=\figheight]{\rfig{S1A_IW_GRDH_1SDV_20170802T163350_34TCR_74_25}{DUCH}{5}}} 
    \end{minipage}
\begin{minipage}[t]{\figwidth}
        \centering
        \centerline{6\textsuperscript{th}}\medskip
        \centerline{\includegraphics[height=\figheight]{\rfig{S1A_IW_GRDH_1SDV_20170802T163350_34TCR_74_25}{DUCH}{6}}} 
    \end{minipage}
     \begin{minipage}[t]{\figwidth}
        \centering
        \centerline{7\textsuperscript{th}}\medskip
        \centerline{\includegraphics[height=\figheight]{\rfig{S1A_IW_GRDH_1SDV_20170802T163350_34TCR_74_25}{DUCH}{7}}} 
    \end{minipage}
     \begin{minipage}[t]{\figwidth}
        \centering
        \centerline{\includegraphics[height=\figheight]{\qfig{S1A_IW_GRDH_1SDV_20170802T163350_34TCR_74_25}}}
        \vspace{-0.05in}\centerline{(a)}\medskip\vspace{-0.06in}
    \end{minipage}
          \begin{minipage}[t]{\figwidth}
        \centering
        \centerline{\includegraphics[height=\figheight]{\rfig{S1A_IW_GRDH_1SDV_20170802T163350_34TCR_74_25}{DSCMR}{1}}} 
    \end{minipage}    
          \begin{minipage}[t]{\figwidth}
        \centering
        \centerline{\includegraphics[height=\figheight]{\rfig{S1A_IW_GRDH_1SDV_20170802T163350_34TCR_74_25}{DSCMR}{2}}} 
    \end{minipage}
          \begin{minipage}[t]{\figwidth}
        \centering
        \centerline{\includegraphics[height=\figheight]{\rfig{S1A_IW_GRDH_1SDV_20170802T163350_34TCR_74_25}{DSCMR}{3}}} 
    \end{minipage}
          \begin{minipage}[t]{\figwidth}
        \centering
        \centerline{\includegraphics[height=\figheight]{\rfig{S1A_IW_GRDH_1SDV_20170802T163350_34TCR_74_25}{DSCMR}{4}}} 
        \vspace{-0.05in}\centerline{(c)}\medskip\vspace{-0.06in}
    \end{minipage}
          \begin{minipage}[t]{\figwidth}
        \centering
        \centerline{\includegraphics[height=\figheight]{\rfig{S1A_IW_GRDH_1SDV_20170802T163350_34TCR_74_25}{DSCMR}{5}}} 
    \end{minipage}
          \begin{minipage}[t]{\figwidth}
        \centering
        \centerline{\includegraphics[height=\figheight]{\rfig{S1A_IW_GRDH_1SDV_20170802T163350_34TCR_74_25}{DSCMR}{6}}} 
    \end{minipage} 
          \begin{minipage}[t]{\figwidth}
        \centering
        \centerline{\includegraphics[height=\figheight]{\rfig{S1A_IW_GRDH_1SDV_20170802T163350_34TCR_74_25}{DSCMR}{7}}} 
    \end{minipage}
     \begin{minipage}[t]{\figwidth}
        \centering
        \centerline{\includegraphics[height=\figheight]{\efig}}
    \end{minipage}
          \begin{minipage}[t]{\figwidth}
        \centering
        \centerline{\includegraphics[height=\figheight]{\rfig{S1A_IW_GRDH_1SDV_20170802T163350_34TCR_74_25}{SSCMCBIRv2}{1}}} 
    \end{minipage}    
          \begin{minipage}[t]{\figwidth}
        \centering
        \centerline{\includegraphics[height=\figheight]{\rfig{S1A_IW_GRDH_1SDV_20170802T163350_34TCR_74_25}{SSCMCBIRv2}{2}}} 
    \end{minipage}
          \begin{minipage}[t]{\figwidth}
        \centering
        \centerline{\includegraphics[height=\figheight]{\rfig{S1A_IW_GRDH_1SDV_20170802T163350_34TCR_74_25}{SSCMCBIRv2}{3}}} 
    \end{minipage}
          \begin{minipage}[t]{\figwidth}
        \centering
        \centerline{\includegraphics[height=\figheight]{\rfig{S1A_IW_GRDH_1SDV_20170802T163350_34TCR_74_25}{SSCMCBIRv2}{4}}} 
        \vspace{-0.05in}\centerline{(d)}\medskip\vspace{-0.06in}
    \end{minipage}
          \begin{minipage}[t]{\figwidth}
        \centering
        \centerline{\includegraphics[height=\figheight]{\rfig{S1A_IW_GRDH_1SDV_20170802T163350_34TCR_74_25}{SSCMCBIRv2}{5}}} 
    \end{minipage}
          \begin{minipage}[t]{\figwidth}
        \centering
        \centerline{\includegraphics[height=\figheight]{\rfig{S1A_IW_GRDH_1SDV_20170802T163350_34TCR_74_25}{SSCMCBIRv2}{6}}} 
    \end{minipage} 
          \begin{minipage}[t]{\figwidth}
        \centering
        \centerline{\includegraphics[height=\figheight]{\rfig{S1A_IW_GRDH_1SDV_20170802T163350_34TCR_74_25}{SSCMCBIRv2}{7}}} 
    \end{minipage}
    \vspace{-0.3cm}
    \caption{S1$\to$S2 retrieval results for (a) S1 query image; and S2 images retrieved by (b) DUCH, (c) DSCMR and (d) our method.}
    \label{fig:soa_comp_visres_figure_DLRSD}
\end{figure*}

%% file: exp_result_table.tex
\begin{table}[t] 
\renewcommand{\arraystretch}{1.2}
\setlength\tabcolsep{0.1pt}
\small
\centering
\caption{Normalized discounted cumulative gain (NDCG) and $F_1$ scores on S1$\to$S2 and S2$\to$S1 retrieval tasks. Requirement of labels (RoL) indicates whether the considered methods require the labeled training images or not.}
\newcommand{\methodCol}{0.1\textwidth}
\newcommand{\typeCol}{0.05\textwidth}
\newcommand{\resCol}{0.055\textwidth}
\label{table:exp_result_table}
\small
\vspace{-0.05in}
\begin{tabular}{
@{}*{1}{>{\arraybackslash}p{\methodCol}}*{1}{>{\centering\arraybackslash}p{\typeCol}}*{6}{>{\centering\arraybackslash}p{\resCol}}} 
\\\Cline{0.9pt}{1-8}
\multirow{2}{\methodCol}[2pt]{\textbf{Method}} & \multirow{2}{\typeCol}[2pt]{\centering\textbf{RoL}}& \multicolumn{2}{c}{\centering\textbf{S1}\boldmath{$\to$}\textbf{S2}} &
\multicolumn{2}{c}{\centering\textbf{S2}\boldmath{$\to$}\textbf{S1}} & \multicolumn{2}{c}{\textbf{Average}} \\\Cline{0.5pt}{3-8}
 & & $F_1$ & NDCG & $F_1$ & NDCG & $F_1$ & NDCG \\\Cline{0.9pt}{1-8}
 S2MC~\cite{Li:2020} & \cmark & 41.7 & 28.0 & 45.6 & 21.1 & 43.7 & 24.5  \\\Cline{0.5pt}{1-8}
 deep-SM~\cite{Wei:2017} & \cmark & 65.0 & 54.1. & 68.7 & 53.8 & 66.8 & 53.9  \\\Cline{0.5pt}{1-8}
  DSCMR~\cite{Zhen:2019} & \cmark & 73.9 & 58.9 & 71.3 & 61.7 & 72.6 & 60.3  \\\Cline{0.5pt}{1-8}
 DCCA~\cite{Andrew:2013} & \xmark & 49.1 & 36.7 & 40.3 & 30.4 & 44.7 & 33.6  \\\Cline{0.5pt}{1-8}
 SimCLR~\cite{Chen:2019} & \xmark & 47.3 & 35.8 & 53.5 & 39.5. & 50.4 & 37.6  \\\Cline{0.5pt}{1-8} 
 DUCH~\cite{Mikriukov:2022} & \xmark & 66.6 & 56.3 & 67.8 & 54.8 & 67.2 & 55.5  \\\Cline{0.5pt}{1-8}
 Ours & \xmark & 71.5 & 59.4 & 71.3 & 58.4 & 71.4 & 58.9 \\\Cline{0.9pt}{1-8} 
\end{tabular}
\vspace{-0.1in}
\end{table}